\documentclass[conference]{IEEEtran}
\IEEEoverridecommandlockouts

\usepackage{cite}
\usepackage{amsmath,amssymb,amsfonts}
\usepackage{algorithmic}
\usepackage{graphicx}
\usepackage{textcomp}
\usepackage{xcolor}
\usepackage{url}
\usepackage{booktabs}
\usepackage{graphicx} 
\usepackage{multirow}
\usepackage{tabularx}

\def\BibTeX{{\rm B\kern-.05em{\sc i\kern-.025em b}\kern-.08em
    T\kern-.1667em\lower.7ex\hbox{E}\kern-.125emX}}
\begin{document}
\title{FreqMark: Frequency-Based Watermark for Sentence-Level Detection of LLM-Generated Text}

\author{
\IEEEauthorblockN{Zhenyu Xu\IEEEauthorrefmark{1}, Kun Zhang\IEEEauthorrefmark{2}, Victor S. Sheng\IEEEauthorrefmark{1}}
\IEEEauthorblockA{\IEEEauthorrefmark{1}\textit{Department of Computer Science}, Texas Tech University, Lubbock, TX, USA \\
\{zhenxu, victor.sheng\}@ttu.edu}
\IEEEauthorblockA{\IEEEauthorrefmark{2}\textit{Department of Computer Science}, Xavier University of Louisiana, New Orleans, LA, USA \\
kzhang@xula.edu}
}

\maketitle

\begin{abstract}
The increasing use of Large Language Models (LLMs) for generating highly coherent and contextually relevant text introduces new risks, including misuse for unethical purposes such as disinformation or academic dishonesty. To address these challenges, we propose FreqMark, a novel watermarking technique that embeds detectable frequency-based watermarks in LLM-generated text during the token sampling process. The method leverages periodic signals to guide token selection, creating a watermark that can be detected with Short-Time Fourier Transform (STFT) analysis. This approach enables accurate identification of LLM-generated content, even in mixed-text scenarios with both human-authored and LLM-generated segments. Our experiments demonstrate the robustness and precision of FreqMark, showing strong detection capabilities against various attack scenarios such as paraphrasing and token substitution. Results show that FreqMark achieves an AUC improvement of up to 0.98, significantly outperforming existing detection methods.
\end{abstract}


\section{Introduction}

The development of Large Language Models has significantly advanced the capabilities of natural language processing, enabling the generation of highly coherent and contextually relevant text. However, this advancement also presents substantial challenges, particularly concerning the misuse of LLM-generated text. Such misuse poses serious threats to AI safety and the integrity of Trustworthy AI initiatives. For instance, LLMs can be exploited to produce false information or fake news, misleading readers and potentially causing public harm \cite{pu}. Additionally, these models are sometimes used for unethical practices such as completing academic assignments for students, which undermines educational integrity \cite{my}. 

Numerous techniques for detecting and watermarking LLM-generated text have emerged \cite{tang}, yet the majority of existing approaches focus primarily on binary classification tasks—distinguishing between human-authored and LLM-generated texts. However, there is a growing need for more precise watermarking technologies and fine-grained detection capable of addressing mixed-text scenarios, where content authored by humans is intermingled with that generated by LLMs. Particularly challenging are attack methods involving mixed texts, such as the insertion of human-authored segments into LLM-generated texts or the paraphrasing of parts of LLM-generated content. These complexities require advanced methods that go beyond simple binary classification to ensure the integrity and origin of text in diverse applications.

To deal with this challenge, we propose a novel method of embedding textual watermarks during the sampling process of LLMs, guided by the sampling of periodic signals. This watermarking technique systematically selects the next token to generate text based on a predefined pattern governed by these periodic signals. By doing so, it embeds a detectable and robust watermark that can be used to trace the origin of the generated text. For mixed-text scenarios, we employ Short-Time Fourier Transform \cite{stft} analysis to precisely target the embedded periodic signals within the watermarked text. This method enables accurate detection of the watermark's presence, distinguishing between human-written and model-generated text. The precision of this technique is maintained within token level. Figure 1 shows an example of concatenated text generated by a copy-paste attack, which consists of a human-written prompt concatenated with text generated by LLM that includes an embedded watermark.

\begin{figure*}[!ht]
    \centering
    \includegraphics[width=0.9\linewidth]{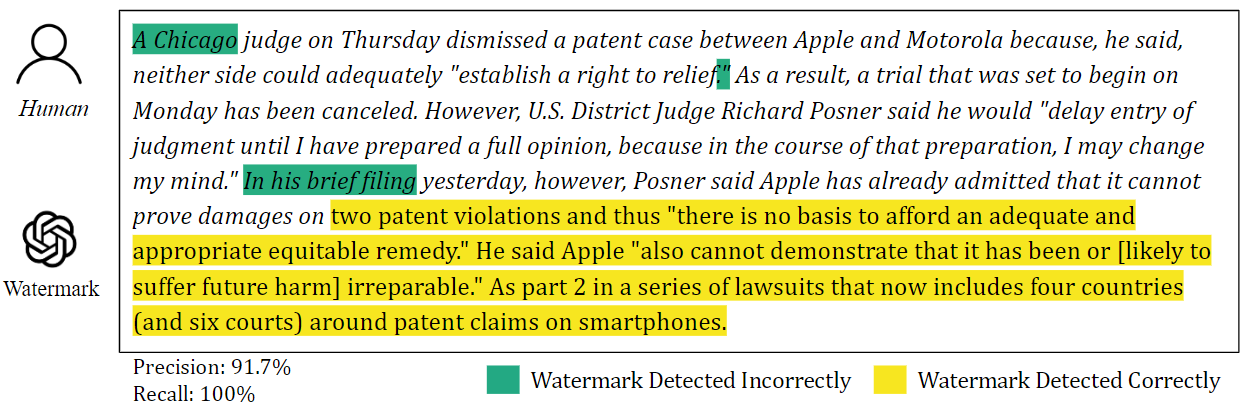}
    \caption{Example of Watermark Detection for Concatenated Text. The slanting typeface text represent human-authored prompt, while the regular typeface parts signify text generated by gpt-3.5-turbo-instruct that embeds watermarks. The yellow highlighted segments indicate portions of the text where the watermark was detected and correctly identified as LLM-generated. Conversely, the green segments represent errors where the text was mistakenly flagged. GPTZero \cite{gptzero} detection completely fails in this sentence, which was highly confident to determine text is entirely human.}
    \label{fig:1}
\end{figure*}

Our contributions are as follows: \begin{itemize} \item We introduce a novel method for embedding soft watermarks during the LLM sampling process, guided by periodic signals, achieving strong watermark robustness. \item We leverage the STFT to detect embedded watermarks, allowing for fine-grained, sentence-level identification of LLM-generated text, even in adversarial conditions like paraphrasing and token substitution. \item We construct a specialized mixed-content dataset based on the realnewslike corpus to rigorously evaluate the fine-grained localization capabilities of watermark detection, enabling comprehensive performance benchmarking under various text manipulation scenarios. \end{itemize}

\section{Related Work}
In recent years, various studies have focused on protecting and verifying the authenticity of LLMs using watermarking techniques to safeguard intellectual property and prevent misuse. Yoo et al. developed a robust multi-bit watermarking system that embeds watermarks into semantically or syntactically significant text components \cite{yoo2023}. Zhao et al. introduced Ginsew, which embeds secret signals into the probability vectors during token generation \cite{zhao2023}. Kirchenbauer et al. proposed a watermarking framework for proprietary language models using a randomized set of "green" tokens \cite{kirchenbauer2023}. Peng et al. created EmbMarker for Embedding as a Service (EaaS) models \cite{peng2023}.

\section{Approach}
\subsection{Watermark Embedding during the LLM Sampling Process}
\subsubsection{Next Token Prediction Process}

Large Language Models generate text by autoregressively predicting tokens based on preceding context. Let \( \mathcal{V} \) represent the set of all tokens (vocabulary), and let \( w \in \mathcal{V} \) denote a specific token. Given a sequence of tokens \( w_{1:(t-1)} = w_1, w_2, \ldots, w_{t-1} \), the model computes a probability distribution for the next token \( w_t \):

\[
P(w_t | w_{1:(t-1)}) = \text{softmax}(f(w_{1:(t-1)}))
\]

where \( f(w_{1:(t-1)}) \) represents the logits for each token based on the given context, and the softmax function \cite{goodfellow2016} is defined as:

\[
\text{softmax}(z_i) = \frac{e^{z_i}}{\sum_{j} e^{z_j}}
\]

\subsubsection{Watermark Embedding Method}

To embed a watermark during the LLM sampling process, we generate logits for each token based on the preceding context and rank these tokens according to their log probability. A periodic guiding signal with predefined amplitude values, such as ranging from 1 to 5, then dictates which rank to select for the next token. The watermark embedding process is compatible with common token sampling strategies, such as temperature scaling and top-k sampling. For example, if the periodic signal is \( \{1, 3, 5, 3, 1\} \), the signal guides the selection of the top-ranked token in the first step, the third-ranked token in the second step, and so on. The watermark embedding is achieved by selecting the next token according to the ranks dictated by the periodic signal. This process repeats cyclically, embedding a watermark in the generated text based on the token ranks. The rank of token \( w_t \), denoted \( \mathcal{R}(w_t) \), is selected such that:
\[
\mathcal{R}(w_t) = a_{(t\mod k)}
\]

where \( k \) is the period of the guiding signal and \( t \) is the token position. The resulting text embeds a detectable frequency pattern, identifiable through signal analysis.

\subsection{Watermark Detection and Analysis}
\subsubsection{Reconstruction of Logits Distribution}

To detect the embedded watermark, we first reconstruct the logits distribution for each token. We assume the LLM sampling process is not mixed with methods like beam search \cite{holtzman2020} or nucleus sampling \cite{holtzman2020}, allowing for a close approximation of the original logits distribution. This assumption holds when the text is fully controlled by our watermarking method. If mixed sampling methods are used, the accuracy of detection may decrease as the reconstructed logits may no longer align with the original. By comparing the computed logits with the tokens, we derive the token ranks, forming a waveform corresponding to the embedded watermark.

\subsubsection{STFT Analysis for Watermark Detection}
\begin{figure}[!ht]
    \centering
    \includegraphics[width=\linewidth]{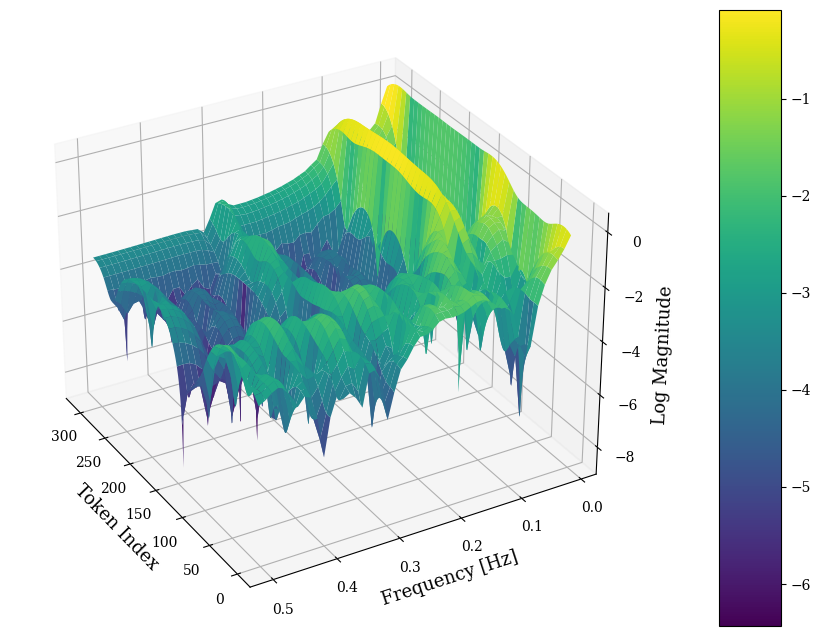}
    \caption{STFT Analysis of Concatenated LLM-generated and Human-authored Text. The bright yellow regions indicate significant frequency components around 0.1 Hz, corresponding to the periodic signal used for watermark embedding. These peaks highlight LLM-generated content, distinguishing it from human-written segments.}
    \label{fig:2}
\end{figure}

To detect the watermark embedded in the generated text, we employ STFT analysis. This signal processing technique enables the examination of the frequency characteristics of the rank sequence extracted from the text. By implementing a sliding window approach, we perform the Fast Fourier Transform (FFT) \cite{fft} within each window to capture the localized frequency content. The sliding window allows for sentence-level detection by analyzing the frequency components within individual segments of text, making it possible to pinpoint LLM-generated content at the sentence level. We apply STFT to the retrieved sequence of rank information from the watermark signal. Figure \ref{fig:2} illustrates an example of STFT analysis applied to a concatenated text, which combines human-written and LLM-generated content.

\section{Experimental Evaluation}
\subsection{Experimental Setting}

We generated text using default GPT-3.5-turbo-instruct \cite{gpt35} by calling the \url{openai.Completion.create} API and additional Llama-2-7b \cite{llama}, both with parameters logprobs=5, temperature=0, and top-p=0.95. We also used GPT-3.5-turbo-instruct to infer the logits distribution. To calculate the perplexity, we used the EleutherAI/gpt-neox-20b \cite{gptneox} model from HuggingFace \cite{hf}. For watermarking, we applied a sinusoidal signal with 10 sampling points per cycle and an amplitude range of 1 to 5, where 10 tokens represent one cycle. The Short-Time Fourier Transform was performed with a window length of 10 points and an overlap of 9, using a Hann window \cite{hann}. We chose a temperature of 0 to eliminate randomness, ensuring the watermark embedding follows the periodic signal precisely. Although compatible with sampling strategies like top-k and temperature scaling, we avoid mixing methods to maintain accurate watermark evaluation. The amplitude and sampling points were selected based on experiments to balance text quality and watermark strength.

\subsection{Dataset Construction}
We constructed a mixed-realnewslike dataset \cite{dataset}, following Kirchenbauer et al., to simulate human edits on LLM-generated text. The dataset covers four attack scenarios: (1) copy-pasting, where human prompts are combined with watermarked text, (2) paraphrasing, where sentences are rewritten while preserving meaning, (3) substitution, using google-t5/t5-3b \cite{t5} to replace token spans, and (4) translation, where text is translated to Spanish and back to English.

\subsection{Baselines}
We used common benchmarks for LLM-generated text detection. Log $p(x)$, entropy, rank, and logrank are based on Kirchenbauer's watermarking experiments. DetectGPT \cite{detectgpt}, a state-of-the-art method, and GPTZero \cite{gptzero}, a commercial detector, were included to ensure a thorough evaluation. DetectLLM \cite{detectllm} introduced LRR (log likelihood and rank) and NPR (an enhancement of DetectGPT). GPT-2 Detector \cite{gpt2detector}, using a RoBERTa-based model \cite{roberta}, was also included. By comparing with both research-oriented methods like DetectGPT and practical tools like GPTZero, we ensure our evaluation is applicable to real-world detection scenarios.

\subsection{Evaluation Metrics}
For watermark detection, we measured False Negative Rate (FNR) and False Positive Rate (FPR) to evaluate how often watermarked and non-watermarked texts were misclassified, and Area Under the Curve (AUC) to assess the model’s ability to distinguish between them. We also used Precision, Recall, and F1 Score for sentence-level detection, where Precision is the proportion of correctly identified LLM-generated tokens, Recall is the proportion of all actual LLM-generated tokens identified, and F1 Score is the harmonic mean of Precision and Recall, providing a balanced measure of performance.

\section{Experimental Results}
\subsection{Basic Watermark Performance}
To demonstrate the effectiveness of our watermarking technique, we conducted sample-level detection tests on a balanced dataset comprising an equal ratio of pure human-written text and pure watermarked text. This evaluation provides a fundamental assessment of the watermark's detectability. Table 1 presents a comparison of FreqMark against various baseline methods, including traditional statistical approaches and state-of-the-art detection tools.

\begin{table}[!ht]
    \centering
    \caption{Comparison of sample-level LLM-generated text detection methods.}
    \begin{tabular}{lccc|ccc}
    \toprule
    \textbf{Methods} & \multicolumn{3}{c}{\textbf{GPT-3.5-turbo-instruct}} & \multicolumn{3}{c}{\textbf{Llama-2-7b}} \\
    \cmidrule(lr){2-4} \cmidrule(lr){5-7}
                     & \textbf{AUC} & \textbf{FPR} & \textbf{FNR} & \textbf{AUC} & \textbf{FPR} & \textbf{FNR} \\
    \midrule
    Log $p(x)$   & 0.52 & 0.35 & 0.30 & 0.54 & 0.38 & 0.32 \\
    Entropy      & 0.48 & 0.38 & 0.52 & 0.62 & 0.30 & 0.58 \\
    Rank         & 0.57 & 0.40 & 0.43 & 0.59 & 0.38 & 0.50 \\
    LogRank      & 0.69 & 0.12 & 0.60 & 0.64 & 0.22 & 0.45 \\
    DetectGPT    & 0.87 & 0.05 & 0.52 & 0.83 & 0.12 & 0.60 \\
    LRR          & 0.66 & 0.35 & 0.33 & 0.70 & 0.30 & 0.35 \\
    NPR          & 0.61 & 0.38 & 0.36 & 0.64 & 0.42 & 0.33 \\
    GPT-2 Detector & 0.51 & 0.87 & 0.19 & 0.45 & 0.80 & 0.25 \\
    \textbf{FreqMark} & 0.98 & 0.08 & 0.02 & 0.96 & 0.10 & 0.04 \\
    \bottomrule
    \end{tabular}
\end{table}

\subsection{Sentence-Level Watermark Detection}

To evaluate our method's ability to distinguish between LLM-generated and human-written content at a fine-grained level, we used an enhanced copy-pasting subset of mixed-content samples. Each sample contains a mix of LLM-generated and human-written text segments, ranging from 20 to 100 tokens, with a minimum 1:1 ratio. The dataset was designed to mimic AI-assisted writing scenarios where human and LLM-generated content are interspersed. Every token in the dataset was labeled as either LLM-generated or human-written. Baseline methods like Log $p(x)$ and GPTZero were selected for their relevance to detecting LLM-generated text in mixed-content scenarios. These methods provide a good comparison as they represent both traditional statistical methods and commercial tools. As shown in Table 2, FreqMark significantly outperforms baselines across all metrics.

\begin{table}[!ht]
\centering
\caption{Comparison of sentence-level LLM-generated text detection methods.}
\begin{tabular}{lccc}
\toprule
\textbf{Methods} & \textbf{Precision} & \textbf{Recall} & \textbf{F1 Score} \\
\midrule
Logp(x)   & 0.16 & 0.18 & 0.17 \\
GPTZero   & 0.21 & 0.15 &  0.18\\
\textbf{FreqMark} & 0.92 & 0.98 & 0.95 \\
\bottomrule
\end{tabular}
\label{table2}
\end{table}

\subsection{Watermark Attack} 

\begin{figure}[!ht]
    \centering
    \includegraphics[width=\linewidth]{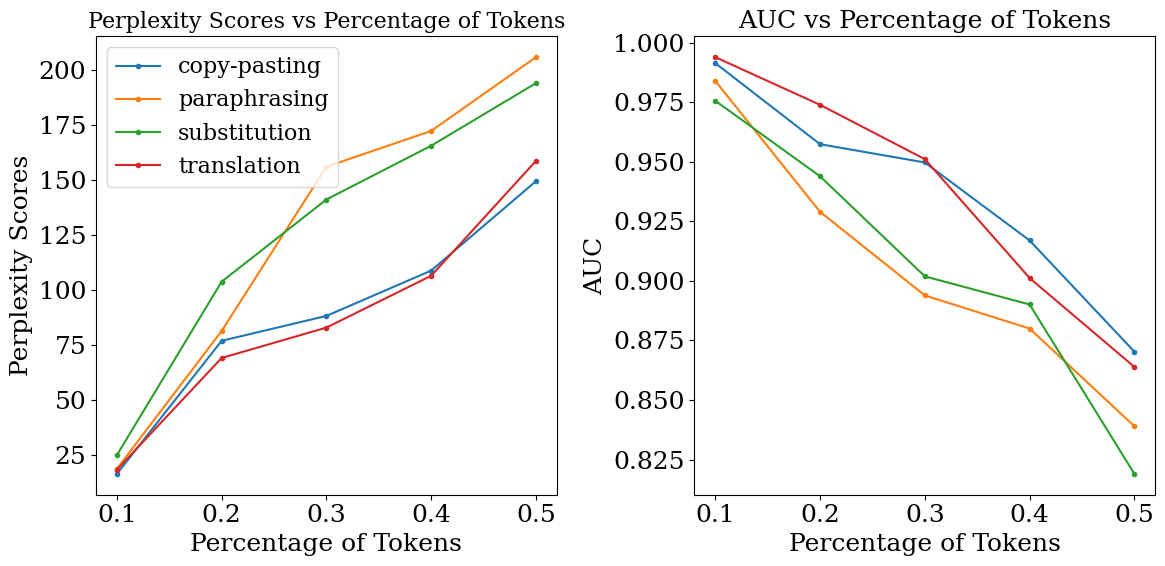}
    \caption{FreqMark watermark robustness under various attack scenarios.}
    \label{fig:3}
\end{figure}

We tested the robustness of the watermark under four attack scenarios: copy-pasting, paraphrasing, substitution, and translation, using the mixed-realnewslike dataset. Paraphrasing and substitution had the largest impact on both perplexity and AUC, likely due to changes in sentence structure disrupting the watermark. In contrast, copy-pasting and translation had less effect. As shown in Figure \ref{fig:3}, the x-axis represents the percentage of token modification, while the left y-axis shows the corresponding Perplexity Scores (indicating text quality), and the right y-axis displays the AUC (measuring the watermark's detectability). 

The results demonstrate that as token modification increases, text quality steadily decreases, particularly in paraphrasing and substitution scenarios, while the watermark remains highly detectable (AUC $\geq$ 0.9) up to about 30-40\% token modification. This threshold reflects typical real-world text manipulations. The periodic signal's resilience allows for reliable detection even with significant text alterations, much like Fourier Transform reveals patterns in noisy data.

\section{Discussion}
\subsection{Limitations and Future Work}
The method may exhibit reduced performance in very short text sequences, where the periodic signal does not have enough tokens to manifest a detectable pattern. Additionally, in cases of extreme text transformations, such as aggressive paraphrasing, heavy token substitutions, or highly structured texts with limited variability, the watermark may become less distinguishable. Future work will explore these edge cases and further improve the method's robustness. Furthermore, in cases of extremely short texts or aggressive paraphrasing, the watermark detection's effectiveness may decrease. Continued research will focus on enhancing the robustness of the watermark in these challenging scenarios.

\subsection{Generalization across LLM Architectures}
The proposed watermarking technique generalizes well across various LLM architectures, as token candidates and logits are standard outputs in most models. This allows our method to be applied not only to models like GPT3.5 but also to a wide range of other LLMs. Ensuring compatibility across different architectures enhances the technique's applicability in diverse environments.

\subsection{Threats to Validity}
Since we use the LLM to predict and calculate the logits distribution across each token in the given text, the preceding text is considered as prompt and significantly influences the probability distribution of subsequent paragraphs. If the text's semantic content is entirely unrelated or spans a large contextual gap, the performance of the watermark detection will be adversely affected.

\section{Conclusion}
This work introduces FreqMark, a frequency-based watermarking technique designed to embed soft watermarks during the LLM text generation process. By guiding token selection with periodic signals, we achieve robust watermark embedding. The STFT analysis allows for precise detection of watermarked content at the sentence level, even under adversarial attacks such as paraphrasing and token substitution. Experimental results show that FreqMark outperforms existing methods in both detection accuracy and robustness. This technique offers a promising solution for mitigating the misuse of LLM-generated content, advancing AI safety and promoting the development of trustworthy AI systems.

\section{Acknowledgments}

We disclose the use of artificial intelligence in this research. The regular typeface text in Figure 1 was generated using OpenAI's GPT-3.5-turbo-instruct model. This AI-generated content serves as a sample to demonstrate our watermarking technique on LLM output. 

This research was supported by the National Institute on Minority Health and Health Disparities (NIMHD) of the National Institutes of Health (NIH) under Award Number U54MD007595.
\vfill\pagebreak



\bibliographystyle{IEEEbib}

\end{document}